\documentclass[10pt,twocolumn,letterpaper]{article}

\usepackage{cvpr}
\usepackage{times}
\usepackage{epsfig}
\usepackage{graphicx}
\usepackage{amsmath}
\usepackage{amssymb}

\usepackage[font=footnotesize]{subfig}
\usepackage[linesnumbered,ruled,boxed]{algorithm2e}

\usepackage[dvipdfmx,pagebackref=true,breaklinks=true,colorlinks,bookmarks=false]{hyperref}

\cvprfinalcopy 

\def\cvprPaperID{1068} 

\ifcvprfinal\fi
\begin{document}

\newcommand{\futo}[1]{\boldsymbol{#1}}
\newcommand{\lw}[1]{\smash{\lower2.0ex\hbox{#1}}}
\newcommand{\lwm}[1]{\smash{\lower1.5ex\hbox{#1}}}
\newcommand{\ope}[1]{\mathop{\rm #1}}
\newcommand{\argmax}{\mathop{\rm argmax}\limits}
\newcommand{\argmin}{\mathop{\rm argmin}\limits}
\newcommand{\red}[1]{\textcolor{red}{#1}}
\newcommand{\blue}[1]{\textcolor{blue}{#1}}
\newcommand{\green}[1]{\textcolor{green}{#1}}

\newcommand{\SSmark}{$\triangledown$}
\newcommand{\Pmark}{$\star$}

\newtheorem{definition}{Definition}

\title{Scalable Solution for Approximate Nearest Subspace Search}

\author{Masakazu Iwamura, Masataka Konishi and Koichi Kise\\
Graduate School of Engineering,
Osaka Prefecture University\\
\{masa, kise\}@cs.osakafu-u.ac.jp}

\maketitle

\begin{abstract}
Finding the nearest subspace is a fundamental problem 
and influential to many applications.
In particular, a scalable solution that is fast and accurate for a
large problem has a great impact.
The existing methods for the problem are, however, useless in
a large-scale problem with a large number of subspaces
and high dimensionality of the feature space.
A cause is that
they are designed based on the traditional idea
to represent a subspace by a single point.
In this paper, we propose a scalable solution for
the approximate nearest subspace search (ANSS) problem.
Intuitively, the proposed method represents
a subspace by multiple points unlike the existing methods.
This makes a large-scale ANSS problem tractable. 
In the experiment with 3036 subspaces in the 1024-dimensional space,
we confirmed that the proposed method was
7.3 times faster than the previous state-of-the-art
without loss of accuracy.
\end{abstract}

\section{Introduction}
{\it Subspace representation},
which represents something by a linear subspace in the Euclidean space,
attracts increasing attention in the computer vision community.
Some examples of research work related to it include
activity recognition~\cite{Turaga_TPAMI2011,Sanin_WACV2013},
video clustering~\cite{Turaga_TPAMI2011},
pedestrian detection~\cite{Hong_ECCV2014},
face recognition~\cite{Wang_PR2006,Turaga_TPAMI2011,Harandi_CVPR2011},
object recognition~\cite{Harandi_CVPR2011,ChenSK_CVPR2013,Cherian_ECCV2014},
feature representation~\cite{Kise_ACPR2011,Wang_ECCV2014},
gender recognition~\cite{Harandi_ECCV2014_kernel}
and
MRI data analysis~\cite{Kim_ECCV2014}.


A major usage of the subspace representation is for pattern recognition,
which requires to find the nearest subspace
to the query subspace.
Thus, this problem, called nearest subspace search (NSS) problem,
is fundamental and influential to many applications.
In particular, a scalable solution that is fast and accurate for a
large problem has a great impact
and is expected to be indispensable in the near future.

The main difficulty of the NSS problem is that
the distance between subspaces is not measured by
a common distance (e.g., the Euclidean distance)
defined in the Euclidean space.
Instead, it is measured by a special kind of distances
defined in the Grassmannian,
where it is regarded as the set of linear subspaces
and a point in the manifold represents a subspace.
Thus, solutions of the well-studied approximate nearest
neighbor search (ANNS) problem%
\footnote{
  To avoid confusion between ANSS and ANNS,
  we add \SSmark (representing a subspace) to ANSS
  and \Pmark (representing a point) to ANNS.
},
which finds the nearest point to the query point,
are not directly applicable to the problem.




To cope with the difficulty, two approaches have been proposed.
One is to develop an approximate nearest subspace search (ANSS\SSmark) method
dedicated to the Grassmannian~\cite{XuWang_ICCV2013}.
The method uses a distance defined in the Grassmannian.
In this approach, the existing framework of the ANNS\Pmark\ 
problem is applied to the ANSS\SSmark\  problem.
However, even with an excellent algorithm,
one cannot avoid the heavy computational burden of
principal component analysis (PCA) or
singular value decomposition (SVD)
required to calculate every distance between subspaces.
The other is to map points in the Grassmannian to
the Euclidean space%
~\cite{Basri_CVPR2007,Basri_ICCVW2009,Basri_TPAMI2011}.
In this approach,
solutions of the ANNS\Pmark\  problem are usable as they are.
However, 
the dimensionality of the mapped space is too high to benefit
from the ANNS\Pmark\  methods.

The existing methods of both approaches do not efficiently solve the problem.
This is because they are developed based on the traditional idea
that \textit{a single instance (such as subspace or point) should be
  represented by a single point},
which has been cultivated through the successful experience in
solving the ANNS\Pmark\  problem.
Indeed, they all represent a subspace by a single point in either
the Grassmannian or Euclidean space
and directly apply the ANNS\Pmark\  to the points.




In this paper, we propose a novel method that is
computationally efficient in a large-scale ANSS\SSmark\  problem.
The proposed method is scalable to both the number of subspaces and
the dimensionality of the feature space.
Its main idea is to decompose a distance calculation in the Grassman
manifold into multiple distance calculations in the Euclidean space.
Each of the distance calculations is efficiently
realized by an existing ANNS\Pmark\  method
in a different usage from the usual.
Thus, this approach can be interpreted to represent
a subspace by multiple points in the Euclidean space.
This makes a large-scale ANSS\SSmark\  problem tractable. 

The contributions of this paper are listed below.
\begin{itemize}
\item The paper presents a scalable solution in the
  ANSS\SSmark\  problem,
  with regard to both the number of subspaces and
  the dimensionality of the feature space.

\item The main contribution of the proposed method
  can be intuitively interpreted to
  represent a subspace by multiple points
  while the existing methods adhere the traditional idea
  to represent a subspace by a single point.


\item  In the experiment with 3036 subspaces in the 1024-dimensional space,
we confirmed that the proposed method was
7.3 times faster than the previous state-of-the-art
without loss of accuracy.
\end{itemize}

\section{Preparation}
\label{sec:preparation}

This section provides necessary knowledge to read the paper
that includes the Grassmannian,
principal angles and distances between subspaces
as well as the relationship between
the Euclidean distance and an inner product.
A part of Secs.~\ref{subsec:Grassmannian},
\ref{subsec:principal_angles},
\ref{subsec:Distances_Between_Subspaces}
is based on \cite{Hamm_ICML2008}.

Throughout the paper,
bold capital letters denote matrices (e.g., $\futo{X}$) and
bold lower-case letters denote column vectors (e.g., $\futo{x}$).

\subsection{Squared Euclidean distance and inner product}
\label{subsec:equivalence}

Let us begin with reviewing
the equivalence of the squared Euclidean distance
and an inner product in a certain condition.

A squared Euclidean distance between two vectors
$\futo{a}$ and $\futo{b}$ is given as
\begin{align}
  \label{eqn:euclidean_distance1}
  d_\text{Euc}^2 (\futo{a}, \futo{b})
  &=
  \| \futo{a} - \futo{b} \|^2 \\
  &=
  \| \futo{a} \|^2
  + \| \futo{b} \|^2
  - 2 \futo{a}^\text{T} \futo{b}.
\end{align}
If $\futo{a}$ and $\futo{b}$ are unit vectors,
i.e., $\| \futo{a} \| = \| \futo{b} \| = 1$,
it becomes
\begin{align}
  \label{eqn:euclidean_distance2}
  d_\text{Euc}^2 (\futo{a}, \futo{b})
  &=
  2 - 2 \futo{a}^\text{T} \futo{b}.
\end{align}
In this case,
the squared Euclidean distance and inner product are
in the relationship of the one-by-one mapping,
where a small distance corresponds to a large inner product
and vice versa.


\subsection{Grassmannian}
\label{subsec:Grassmannian}

\begin{definition}
The Grassmannian $\mathcal{G}(m,D)$ is
the set of $m$-dimensional linear subspaces of the $\mathbb{R}^D$.
\end{definition}

Let $\futo{Y}$ be a $D \times m$ orthonormal matrix such that
$\futo{Y}^{\text{T}} \futo{Y} = \futo{I}_m$,
where $\futo{I}_m$ is the $m \times m$ identity matrix.
$\futo{Y}$ spans a subspace, and
$\ope{span}(\futo{Y})$ denotes the subspace spanned by $\futo{Y}$
\footnote{
The subspace spanned by $\futo{Y}$ is also spanned by other
orthonormal matrices $\futo{Y} \futo{R}$,
where $\futo{R} \in \mathcal{O}(m)$.
$\mathcal{O}(m)$ is the group of $m \times m$ orthonormal matrices.
}.
$\ope{span}(\futo{Y}) \in \mathcal{G}(m,D)$
is regarded as a point in the Grassmannian.

\subsection{Principal angles}
\label{subsec:principal_angles}

To find the nearest subspace,
we have to define a distance between subspaces.
Some distances are calculated
using the \textit{principal angles} defined below.
\begin{definition}
Let $\futo{Y}_1$ and $\futo{Y}_2$ be orthonormal matrices of size
$D \times m$.
The principal angles
$0 \le \theta_1 \le \dots \le \theta_m \le \pi/2$ between two
subspaces $\ope{span}(\futo{Y}_1)$ and $\ope{span}(\futo{Y}_2)$
can be computed from the SVD of $\futo{Y}_1^\text{T} \futo{Y}_2$ as
\begin{align}
  \futo{Y}_1^\text{T} \futo{Y}_2
  &=
  \futo{U} \futo{\Sigma} \futo{V}^{\text{T}},
\end{align}
where $\futo{\Sigma}$ is an
$m \times m$
diagonal matrix such that
$\futo{\Sigma} = \ope{diag} ( \cos \theta_1 \dots, \cos \theta_m)$%
, and
$\futo{U}$ and $\futo{V}$ are
$m \times m$
orthonormal matrices such that
$\futo{U} = [ \futo{u}_1 \  \dots \ \futo{u}_{m} ]$ and
$\futo{V} = [ \futo{v}_1 \ \dots \ \futo{v}_{m} ]$, respectively.
\end{definition}

\subsection{Distances between subspaces}
\label{subsec:Distances_Between_Subspaces}

The \textit{geodesic distance} is a formal measure of the distance
between two subspaces.
It is the length of the shortest geodesic connecting the two points
on the Grassmannian.
Using the principal angles, the geodesic distance is given as
\begin{align}
  \label{eqn:geodesic}
  d_\text{G} ( \futo{Y}_1, \futo{Y}_2 )
  &=
  \left(
  \sum_i \theta_i^2
  \right)^{1/2}.
\end{align}
%
The geodesic distance is computationally expensive because
it requires either PCA or SVD 
to calculate the principal angles.

In the literature~\cite{Hamm_ICML2008,Edelman_SIAM1998},
some other distances are defined using the principal angles.
Among them, we introduce the \textit{projection metric} given below.
\begin{align}
  \label{eqn:projection_metric}
  d_\text{P} (\futo{Y}_1, \futo{Y}_2)
  &=
  \left(
  \sum_{i=1}^m \sin^2 \theta_i
  \right)^{1/2} \\
  &=
  \left(
  m -
  \sum_{i=1}^m \cos^2 \theta_i
  \right)^{1/2}
\end{align}
%
This is one of two metrics kernelized in~\cite{Hamm_ICML2008}.
The kernelized version of the projection metric
called \textit{projection kernel} is given as follows.
\begin{align}
  \label{eqn:projection_kernel}
  k_\text{P} (\futo{Y}_1, \futo{Y}_2)
  &=
  \| \futo{Y}_1^\text{T} \futo{Y}_2 \|_F^2.
\end{align}
Note that the projection kernel represents similarity
while the projection metric does distance.
The projection kernel is computationally cheaper than
the geodesic distance
because it does not depend on the principal angles.
As is described in Sec.~\ref{sec:proposed},
it has a desirable property for the proposed method.

Other kernels for the Grassmannian are proposed in%
~\cite{Harandi_WACV2012,Harandi_ECCV2014_kernel}.
Among them, we introduce
the \textit{Grassmannian radial basis function (Grassmannian RBF; GRBF) kernel}%
~\cite{Harandi_ECCV2014_kernel}
given as
\begin{align}
  \label{eqn:GRBF_kernel}
  k_\text{GRBF} (\futo{Y}_1, \futo{Y}_2)
  &=
  \exp \left(
  \beta \| \futo{Y}_1^\text{T} \futo{Y}_2 \|_F^2
  \right), \ \beta>0.
\end{align}
The Grassmannian RBF kernel also has a desirable property for the
proposed method.


\section{Related Work}
\label{sec:related}

In this section, we review related work
in greater detail than in the introduction
using the terminologies introduced in the previous section.

A representative method to solve the ANSS\SSmark\  problem
is the one proposed by Basri et al.
in a series of researches%
~\cite{Basri_CVPR2007,Basri_ICCVW2009,Basri_TPAMI2011}.
We call the method BHZ, following \cite{XuWang_ICCV2013}.
Their main idea is to map a linear subspace
(i.e., a point in the Grassmannian)
to a point in the Euclidean space
so as to apply an ANNS\Pmark\  method.
As is already pointed out,
its major drawback is
that the dimensionality of the mapped space is too high to benefit
from the ANNS\Pmark\  methods.
Letting $D$ be the dimensionality of the original feature space,
that of the mapped space is $D(D+1)/2$.
For example, subspaces in the 1024- and 256-dimensional spaces
used in the experiments in the paper
are respectively mapped to points in the 524800- and 32896-dimensional spaces.
In such a high-dimensional space,
no ANNS\Pmark\  method efficiently works and
even the brute-force search does not work.

Wang et al. propose another kind of method called
Grassmanian-based locality hashing (GLH).
It realizes the framework of
the locality sensitive hashing (LSH)%
~\cite{Har-Peled_TOC2012} in the Grassmannian
using the geodesic distance.
Its main idea is to index the subspaces in the database
with random vectors.
More precisely,
for each random vector, subspaces are divided into two states
based on whether the angle between a random vector and a subspace
is within a threshold angle
(less than or equal to $\pi/6$) or not.
It is, however, not practical
in the large-scale ANSS\SSmark\  problem.
A cause is that in a high dimensional space,
the angle of two vectors tends to be close to orthogonal;
almost no chance for the angle to be less than or equal to $\pi/6$.
This means that it cannot divide subspaces as desired.
In addition, as already pointed out,
the fact that computationally expensive PCA or SVD is
required to calculate every geodesic distance between subspaces
is a heavy burden.





\section{Proposed Method}
\label{sec:proposed}

\subsection{Problem definition}
\label{subsec:problem_definition}

The problem we address is to find the nearest subspace from
the subspaces stored in the database,
each of which is denoted by $\ope{span}(\futo{P}_i)$,
to the given query subspace,
denoted by $\ope{span}(\futo{Q})$.
Here, $\futo{P}_i$ and $\futo{Q}$ are orthonormal matrices of size
$D \times m$.
The problem is formulated as
\begin{align}
  \label{eqn:objective_func_dist}
  i^*
  &=
  \argmin_{i}
  \ope{dist} (\futo{P}_i, \futo{Q})
\end{align}
or
\begin{align}
  \label{eqn:objective_func_sim}
  i^*
  &=
  \argmax_{i}
  \ope{sim} (\futo{P}_i, \futo{Q}),
\end{align}
where $i^*$ is the ID of the nearest subspace to the query subspace,
and $\ope{dist} (\cdot, \cdot)$ and
$\ope{sim} (\cdot, \cdot)$ are some distance and
similarity functions between subspaces, respectively.
In this paper,
we use the projection kernel given in Eq.~\eqref{eqn:projection_kernel}
and the Grassmannian RBF kernel
given in Eq.~\eqref{eqn:GRBF_kernel}
as the similarity function in Eq.~\eqref{eqn:objective_func_sim}%
~\footnote{One may have a question about the pros and cons
of replacing the geodesic distance with these kernels
with regard to accuracy.
However, it is almost impossible to have a general discussion
on it because the accuracy fully depends on data.
Thus, the accuracies of these kernels
may be worse than that of the geodesic distance
though the experimental results in the paper were opposite.}.
Hereafter, we present the proposed method using
the projection kernel.
The outcome on the projection kernel is
directly applicable to the Grassmannian RBF kernel
because Eq.~\eqref{eqn:GRBF_kernel} is rewritten as
\begin{align}
  \label{eqn:GRBF_kernel_rewrite}
  k_\text{GRBF} (\futo{Y}_1, \futo{Y}_2)
  &=
  \exp \left(
  \beta
  k_\text{P} (\futo{Y}_1, \futo{Y}_2)
  \right), \ \beta>0.
\end{align}

\subsection{Approximation of distance calculation}
\label{subsec:approximation}

As mentioned above, the proposed method calculates
a distance between subspaces based on
multiple distances in the Euclidean space.
This is realized by decomposing the similarity function.

\begin{figure}[tbp]
  \begin{center}

    \subfloat[256-dimensional feature vector of the object recognition task.]{
      \includegraphics[width=.9\hsize]{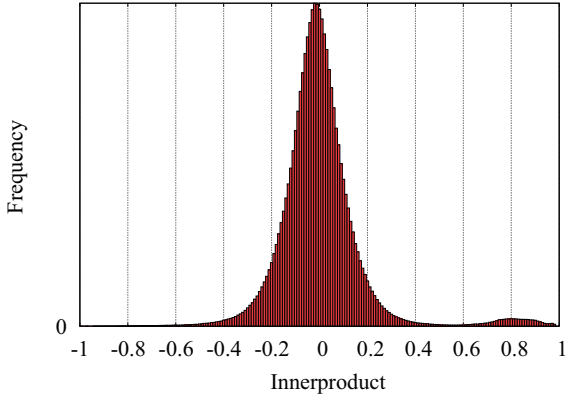}
      \label{fig:histgram_ETH80}
    }

    \subfloat[256-dimensional feature vector of the handwritten character recognition task.]{
      \includegraphics[width=.9\hsize]{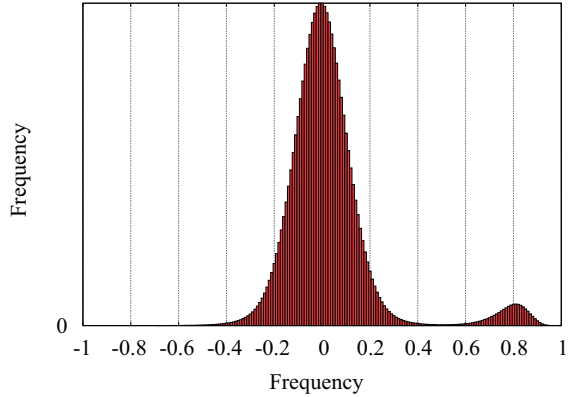}
      \label{fig:histgram_ETL9B_16x16}
    }

    \subfloat[1024-dimensional feature vector of the handwritten character recognition task.]{
      \includegraphics[width=.9\hsize]{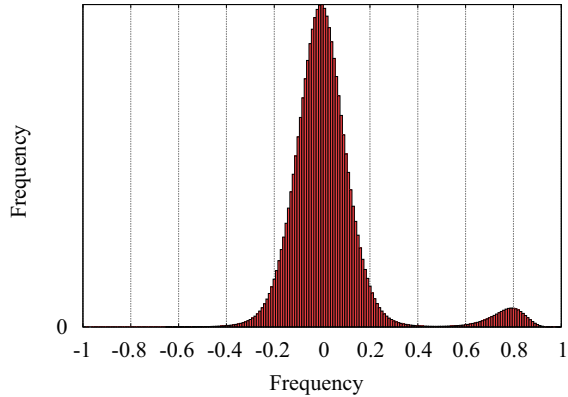}
      \label{fig:histgram_ETL9B_32x32}
    }
    \caption{Histogram of inner products in the object recognition task
      in Sec.~\ref{subsec:ETH80} and
      handwritten character recognition task in Sec.~\ref{subsec:ETL9B}.}
    \label{fig:histogram}

  \end{center}
\end{figure}

Letting
$\futo{P}_i =
[ \futo{p}_{i1} \ \dots \ \futo{p}_{im} ]$
and
$\futo{Q} =
[ \futo{q}_{1} \ \dots \ \futo{q}_{m} ]$,
the elements of their matrix product are given as
\begin{align}
  \label{eqn:Y1Y2_elements}
  \futo{P}_i^\text{T} \futo{Q}
  &=
  \begin{bmatrix}
    \futo{p}_{i1}^\text{T} \futo{q}_{1} &
    \futo{p}_{i1}^\text{T} \futo{q}_{2} &
    \dots &
    \futo{p}_{i1}^\text{T} \futo{q}_{m} \\
%
    \futo{p}_{i2}^\text{T} \futo{q}_{1} &
    \futo{p}_{i2}^\text{T} \futo{q}_{2} &
    \dots &
    \futo{p}_{i2}^\text{T} \futo{q}_{m} \\
%
    \vdots &
    \vdots &
    \ddots &
    \vdots \\
    \futo{p}_{im_1}^\text{T} \futo{q}_{1} &
    \futo{p}_{im_1}^\text{T} \futo{q}_{2} &
    \dots &
    \futo{p}_{im_1}^\text{T} \futo{q}_{m}
  \end{bmatrix}.
\end{align}
%
Thus, Eq.~\eqref{eqn:projection_kernel} of the projection kernel
can be deformed as
\begin{align}
  \label{eqn:projection_kernel_deform}
  k_\text{P} (\futo{P}_i, \futo{Q})
  &=
  \| \futo{P}_i^\text{T} \futo{Q} \|_F^2
  =
  \sum_{s=1}^{m}
  \sum_{t=1}^{m}
  \left(
  \futo{p}_{is}^\text{T} \futo{q}_{t}
  \right)^2.
\end{align}
In contrast to the squared Euclidean distance
having only one inner product
in Eq.~\eqref{eqn:euclidean_distance2},
the projection kernel has
$m^2$
inner products.
Most of them, however, do not contribute to determine
the value of the projection kernel.
Fig.~\ref{fig:histogram} shows typical distributions
of inner products.
%
As seen in the figure, most of the inner products take
values close to zero.
Favorably, in a higher dimensional space,
more inner products take values close to zero.
It is because the sum of each column of
$\futo{P}_i^\text{T} \futo{Q}$ in Eq.~\eqref{eqn:Y1Y2_elements}
is bounded by one regardless of dimensionality.
That is,
for a query eigenvector $\futo{q}_l$,
$  \sum_{t=1}^m \left( \futo{p}_{it}^\text{T} \futo{q}_l \right)^2 \le 1 $.
This can be geometrically interpreted as that
\textit{the length of a unit vector projected on a subspace is less than one}.
This implies that only a limited number of inner products contribute
to determine the value of the projection kernel, and
more importantly they determine
the order of these values of different subspaces.

As easily conceivable from the property,
taking dominating inner products is sufficient to select
the nearest subspace with the maximum similarity.
Here a question arises. How can we \textit{efficiently} take
the dominating (large-valued) inner products?
One might think of a process such as
(1) calculate all inner products first and
then (2) select large-valued ones.
However, this does not help reduce computational time
because the process (1) requires much time while (2) does not.

Our strategy efficiently realizes it by using an ANNS\Pmark\ 
method in a different usage from the usual.
As seen in Eq.~\eqref{eqn:euclidean_distance2},
the squared Euclidean distance and inner product are equivalent
for unit vectors.
Recall that a small distance means a large inner product.
Thus, finding some vectors with large-valued inner products is equivalent
to finding the same number of vectors with small Euclidean distances.
An important note is that
we need a special care to treat \textit{squared values} of inner products
in this scheme.
Since squared values of inner products are used
in Eq.~\eqref{eqn:projection_kernel_deform},
not only vectors having small distances
but also ones having large distances should be retrieved.
So as to cope with this,
query vectors with the opposite signs
(e.g., $-\futo{q}_l$ for $\futo{q}_l$)
are also used as queries of the ANNS\Pmark\  problem.
Since $-\futo{q}_l$ is the most distant vector
to $\futo{q}_l$ on the surface of the unit hyper-sphere,
the most distant vectors to $\futo{q}_l$ are obtained
as the nearest neighbors to $-\futo{q}_l$.

One might worry about robustness of the proposed method
against rotation of bases that span subspaces (i.e., $\futo{P}_i$)
because of the following reason.
\begin{quote}
  The Euclidean distance between two vectors is not preserved
  after the two vectors are differently rotated.
  That is,
  \begin{align}
    \label{eqn:Euclidean_preserve}
    & \| \futo{a} - \futo{b} \|^2
    \ne
    \| \futo{R}_1 \futo{a} - \futo{R}_2 \futo{b} \|^2, \nonumber \\
    & \qquad
    \mathrm{for}\ \futo{R}_1, \futo{R}_2 \in \mathcal{O}(m), \ 
    \futo{R}_1 \ne \futo{R}_2.
  \end{align}
  The proposed method selects large-valued inner products based on
  the Euclidean distance.
  Thus, if the bases are rotated,
  the proposed method cannot correctly select
  the large-valued inner products.
\end{quote}
Though the Euclidean distance is affected by rotation of bases
like Eq.~\eqref{eqn:Euclidean_preserve} and
the proposed method is based on the Euclidean distance,
they do not mean that the proposed method is spoiled by rotation of bases.
The reason is that the proposed method does not try to select the same
inner products regardless of rotation of bases but adaptively selects
large-valued inner products.
Rotation of bases changes the values of elements (inner products)
of the matrix in the right-hand side of
Eq.~\eqref{eqn:Y1Y2_elements}.
Hence, large-valued inner products that
should be selected by the proposed method are changed.
They are adaptively selected by an ANNS\Pmark\  method which can
efficiently select near points to a query
(near points correspond to large-valued inner products).

  

%
\begin{algorithm}[t]
  \caption{Indexing procedure of the proposed method.} 
  \label{alg:indexing}

  \SetCommentSty{textrm}

  \KwIn{Vector data to calculate their subspaces}

  \KwOut{Eigenvectors $\{ \futo{P}_i \}$
    indexed by an ANNS\Pmark\  method}

  \tcp{\bf [Step 1] Calculate eigenvectors}

  Calculate the eigenvectors
  $\futo{P}_i =
  [ \futo{p}_{i1} \ \dots \ \futo{p}_{im} ]$
  of the subspace $i$ to be stored in the database,
  for all $i$.

  \tcp{\bf [Step 2] Store eigenvectors into database}
  Store all the column vectors,
  i.e., $\forall i, \futo{p}_{i1}, \dots, \futo{p}_{im}$,
  into the database.

  \tcp{\bf [Step 3] Indexing} 
  Execute the indexing process of an ANNS\Pmark\  method.
\end{algorithm}
\begin{algorithm}[tbp]
  \caption{Searching procedure of the proposed method.} 
  \label{alg:searching}

  \SetCommentSty{textrm}

  \KwIn{Query vectors to calculate its subspace,
    Eigenvectors $\{ \futo{P}_i \}$
    indexed by the ANNS\Pmark\  method}

  \KwOut{The nearest subspace(s)}

  \tcp{\bf [Step 1] Calculate query eigenvectors}
  Calculate
  $\futo{Q} =
  [ \futo{q}_{1} \ \dots \ \futo{q}_{m} ]$
  in the same manner as in the indexing procedure.

  \tcp{\bf [Step 2] ANNS\Pmark\  search}
  Using the ANNS\Pmark\  method,
  search the $k$ (approximate) nearest neighbors to each of column vectors
  ($\futo{q}_{1}, \dots, \futo{q}_{m}$) of $\futo{Q}$
  and ones with opposite signs
  ($-\futo{q}_{1}, \dots, -\futo{q}_{m}$).

  \tcp{\bf Preparation for Step 3}
  Let $s = \{ +, -\}$ be an indicator to represent
  either $\futo{q}_{l}$ or $-\futo{q}_{l}$;
  ``$s=+$'' for $\futo{q}_{l}$ and ``$s=-$'' for $-\futo{q}_{l}$.

  Let $i(k',l,s)$ and $e(k',l,s)$ be
  the subspace ID ($1, \dots, N_\text{sub}$)
  and eigenvector ID ($1, \dots, m$)
  of the $k'$-th nearest neighbor of
  either $\futo{q}_{l}$ or $-\futo{q}_{l}$,
  switched by $s$.

  The $k$ nearest neighbors of $\futo{q}_{l}$
  and $-\futo{q}_{l}$ are given as
  \begin{multline}
    \futo{p}_{i(1,l,+), e(1,l,+)},
    \futo{p}_{i(1,l,-), e(1,l,-)},
    \dots, \\
    \futo{p}_{i(k,l,+), e(k,l,+)},
    \futo{p}_{i(k,l,-), e(k,l,-)}.
  \end{multline}

  \tcp{\bf [Step 3] Calculate an approximate similarity of the projection kernel}
  Calculate an approximate similarity of the projection kernel
  (Eq.~\eqref{eqn:projection_kernel_deform})
  in an incremental manner described
  in Steps 3-1 and 3-2.

  \tcp{\bf [Step 3-1] Initialize similarities}
  Initialize the similarities of all subspaces with 0 by
  \begin{align}
  \forall i,
  k_\text{P} (\futo{P}_i, \futo{Q})
  \leftarrow 0.
  \end{align}

  \tcp{\bf [Step 3-2] Update similarities}
  For all $k'$, $l$ and $s$,
  update the similarity as follows.
  \begin{align}
    \label{eqn:projection_kernel_increment}
    k_\text{P} (\futo{P}_{i(k',l,s)}, \futo{Q})
    &\leftarrow
    k_\text{P} (\futo{P}_{i(k',l,s)}, \futo{Q}) \\ \nonumber
    & \qquad
    +
    \left(
    \futo{p}_{i(k',l,s), e(k',l,s)}^\text{T} \futo{q}_{l}
    \right)^2.
  \end{align}

  \tcp{\bf [Step 4] Select the nearest subspace(s)}
  Select the subspace(s) having the largest similarity(ies)
  as the nearest subspace(s).
\end{algorithm}

\subsection{Procedure}
\label{subsec:procedure}

The procedure of the proposed method is given as follows.

\subsubsection{Indexing}

The indexing procedure of the proposed method is shown in
Algorithm~\ref{alg:indexing}.
In the process,
all eigenvectors spanning the subspaces are stored in the database.
Then, they are indexed following the manner of an ANNS\Pmark\  method.


\subsubsection{Search}

The searching procedure of the proposed method is shown in
Algorithm~\ref{alg:searching}.
In the process, 
Eq.~\eqref{eqn:projection_kernel_deform} is calculated
in an incremental manner;
larger inner products are summed up earlier.
The computation is approximated by quitting the process
before all inner products are calculated.
If an ANNS\Pmark\  method finds $k$ nearest neighbors to 
each of $2m$ vectors,
The total number of inner products calculated is 
$2km$.
The $2m$ vectors consist of
$m$ query eigenvectors $\{q_i\}$ and
ones with opposite signs to $\{q_i\}$ (i.e., $\{-q_i\}$).
With $k = m N_\text{sub} / 2$,
the proposed method outputs the result without approximation,
where $N_\text{sub}$ is the number of subspaces.


It is noteworthy that
it is not necessary to calculate the inner product
$\futo{p}_{i(k',l,s), e(k',l,s)}^\text{T} \futo{q}_{l}$
in Eq.~\eqref{eqn:projection_kernel_increment}
because it is
provided by the ANNS\Pmark\  method.
That is, from Eq.~\eqref{eqn:euclidean_distance2},
the inner product is given in the form of
\begin{align}
  \label{eqn:innerproduct_from_Euclidean_distance}
  \futo{a}^\text{T} \futo{b}
  &=
  1 -
  \frac{
    d_\text{Euc}^2 (\futo{a}, \futo{b})
  }{2}.
\end{align}
%
This also helps reduce computational time of the proposed method.

\section{Experiments}
\label{sec:experiments}

To evaluate the scalability of the proposed methods,
a large dataset in terms of the number of subspaces is desirable.
As far as the authors know, however,
there is no appropriate dataset that satisfies
both (1) the number of categories (subspaces) is
large (preferably 10000+) and
(2) multiple samples per category are available.
Thus, we used a dataset for object recognition
that is commonly used to evaluate subspace classification methods
and a relatively large dataset
for handwritten Japanese character recognition.


%
\begin{table*}[t]
  \begin{center}
    \caption{Summary of the methods and parameters used in the experiments.
    * in abbreviation indicates the proposed methods.}
    \label{tbl:parameters}
    \begin{tabular}{|c|c|c|c|c|}
      \hline
      \lwm{Type} & \lwm{Abbreviation} & \lwm{Method}
      & Parameters used in & Parameters used in
      \\
      & & & Sec.~\ref{subsec:ETH80}
      & Sec.~\ref{subsec:ETL9B}
      \\ \hline \hline
      & GD & Geodesic distance in Eq.~\eqref{eqn:geodesic}
      & N/A & N/A \\ \cline{2-5}
      NSS & PK & Projection kernel in Eq.~\eqref{eqn:projection_kernel}
      & N/A & N/A \\ \cline{2-5}
      & GRBF & Grassmannian RBF kernel
      & $\beta=1$ & $\beta=1$ \\ \hline \hline
      & BHZ & ANSS\SSmark\  method by Basri et al.~\cite{Basri_TPAMI2011}
      & N/A & N/A \\ \cline{2-5}
      & \lwm{GLH}
      & \lwm{Grassmanian-based locality hashing~\cite{XuWang_ICCV2013}}
      & \multicolumn{2}{|c|}{
        Combinations of $S = 100, 500,1000,5000,10000$}
      \\
      ANSS\SSmark & & & \multicolumn{2}{|c|}{and $K = 1, 2, 3, 4, 5$}
      \\ \cline{2-5}
%
      & APK* & Approximate projection kernel
      & $k = 1, 2, \dots, 280$ & $k = 1, 101, \dots, 2001$ \\ \cline{2-5}
      & AGRBF* & Approximate Grassmannian RBF kernel
      & $k = 1, 2, \dots, 280$ & $k = 1, 101, \dots, 2001$ \\ \hline
    \end{tabular}
  \end{center}
\end{table*}
The summary of the ANSS\SSmark\  methods used in the experiments
is shown in Table~\ref{tbl:parameters}.
As the proposed methods,
in addition to the approximate projection kernel (APK)
presented in Sec.~\ref{sec:proposed},
approximate Grassmannian RBF kernel (AGRBF)
obtained by replacing
the projection kernel (PK) in Eq.~\eqref{eqn:projection_kernel}
with
Grassmannian RBF kernel in Eq.~\eqref{eqn:GRBF_kernel}
was also used.
The proposed methods were compared with
existing NSS and ANSS\SSmark\  methods.
The purpose of comparison with the NSS methods is to evaluate
how much computational time the proposed methods can
save \textit{with a reasonable loss of accuracy}.
As the NSS methods, geodesic distance (GD) in Eq.~\eqref{eqn:geodesic},
projection kernel (PK) in Eq.~\eqref{eqn:projection_kernel},
and Grassmannian RBF kernel (GRBF) in Eq.~\eqref{eqn:GRBF_kernel}
were used.
The purpose of comparison with the ANSS\SSmark\  methods is to evaluate
how much computational time the proposed methods can
save \textit{without loss of accuracy}.
As ANSS\SSmark\  methods, 
the one proposed by
Basri et al. (BHZ)~\cite{Basri_TPAMI2011},
and
Grassmanian-based locality hashing (GLH)~\cite{XuWang_ICCV2013}
were used.

All the methods were implemented in the C++ language by the authors.
The liboctave library was used to efficiently calculate
SVD and PCA as well as matrix products.
As the ANNS\Pmark\  method for the proposed methods
(APK and AGRBF), the bucket distance hashing (BDH)%
~\cite{Iwamura_ICCV2013} was used.
Though we tried to use it also in BHZ,
it did not work due to too high dimensionality.
Thus, we used the brute-force search method instead.
GLH indexes the subspaces in the database
by dividing the angles between the subspaces and random vectors
into two.
We fixed its threshold to $\pi/6$ as suggested in the paper.
For other methods (GD, PK and GRBF),
the brute-force search method was used.
Note that PK and GRBF achieve the same recognition accuracy
because their difference is
whether the exponential function is taken or not,
and the order of similarities of subspaces
do not change in PK and GRBF.
In the same reason,
APK and AGRBF achieve the same recognition accuracy.

We employed servers where
4 CPUs (Intel Xeon E5-4627 v2, 3.3GHz, 8 cores) and
512GB memory were installed.
All data were stored on memory.
Each program was executed as a single thread on a single core.
The average results of seven runs are shown below.

\subsection{Object recognition}
\label{subsec:ETH80}

\begin{figure}[tbp]
  \begin{center}
    \includegraphics[width=\hsize]{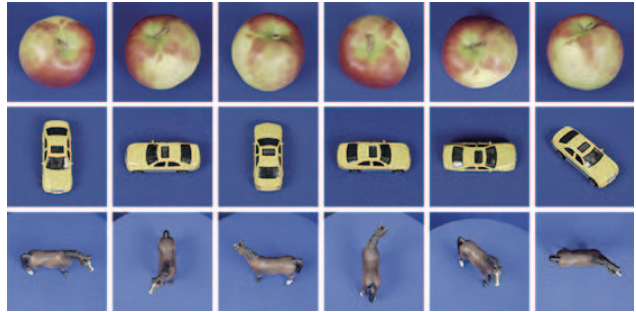}
    \caption{Sample images in the ETH-80 database.}
    \label{fig:eth80}
  \end{center}
\end{figure}

The ETH-80 dataset contained eight object categories,
each of which contained ten objects~\cite{Leibe_CVPR2003}.
Thus, we had 80 categories in total.
For each object,
41 images captured from different viewpoints were available.
Sample images are shown in Fig.~\ref{fig:eth80}.
\begin{figure}[tbp]
  \begin{center}
    \includegraphics[width=\hsize]{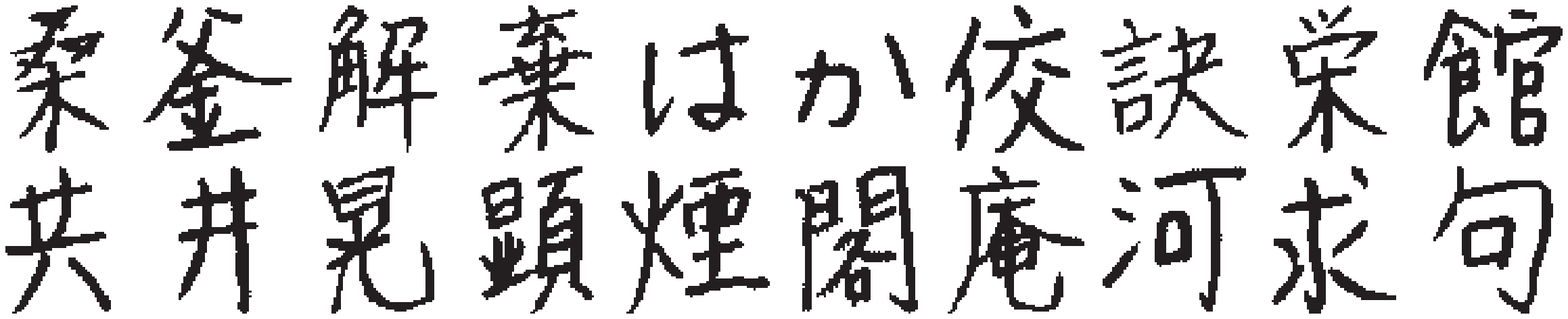}
    \caption{Sample images of the ETL9B database.}
    \label{fig:etl9b}
  \end{center}
\end{figure}
The resolution of the images was $256 \times 256$.
The images were converted to $16 \times 16$ grayscale images,
and each pixel value was used as a feature.
Finally, 256-dimensional feature vectors were obtained.
The images of odd numbers (21 images per category) were used for training and
the ones of even numbers (20 images per category) were used for testing.
The number of subspaces stored in the database was 80
(equivalent to the number of categories) and
that of query subspaces was 880.
880 comes from use of 11 query subspaces for each category.
The reason we had 11 query subspaces was that,
among 20 training images,
consecutive 10 images were selected 11 times.

The dimensionality of subspaces was determined
in the preliminary experiment
to achieve the highest accuracy by PK.
It was $m=7$.
Since the number of subspaces (categories) in the database was 80,
the total number of inner products was 3920.
3920 comes from $80 \times 7 \times 7$.
Since the dimensionality $m$ of subspaces was determined to be 7,
the number of inner products of Eq.~\eqref{eqn:Y1Y2_elements}
became 49 for each category.
Thus, with $k = 280$,
the proposed methods APK and AGRBF achieve the same accuracies as
PK and GRBF as long as all the true $k$-nearest neighbors are
retrieved by the ANNS\Pmark\ method.

\begin{figure}[tbp]
  \begin{center}
    \subfloat[$k$ vs Accuracy]{
      \includegraphics[width=.95\hsize]{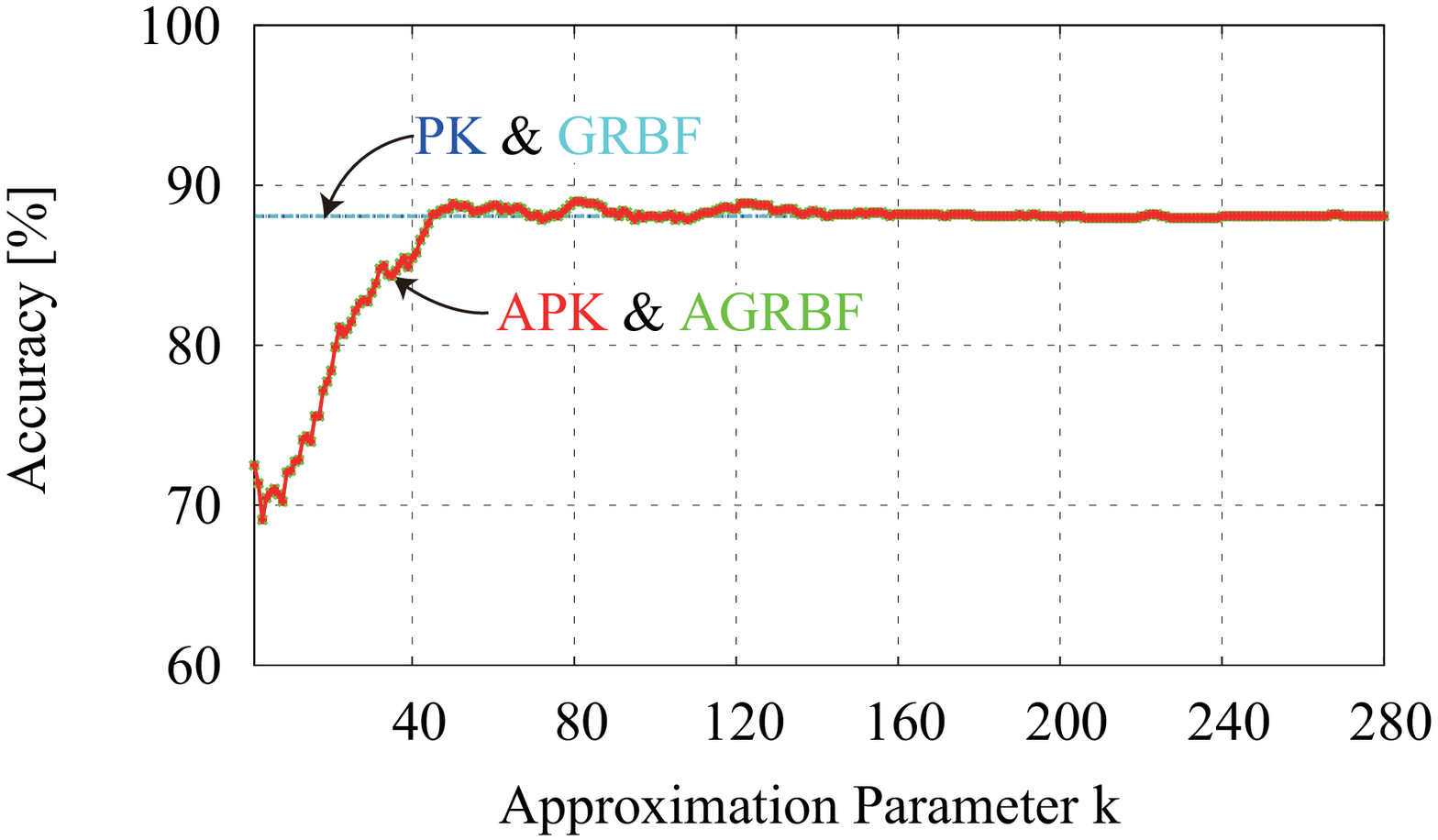}
      \label{fig:ETH80_k_vs_Acc}
    }

    \subfloat[$k$ vs Time]{
      \includegraphics[width=.95\hsize]{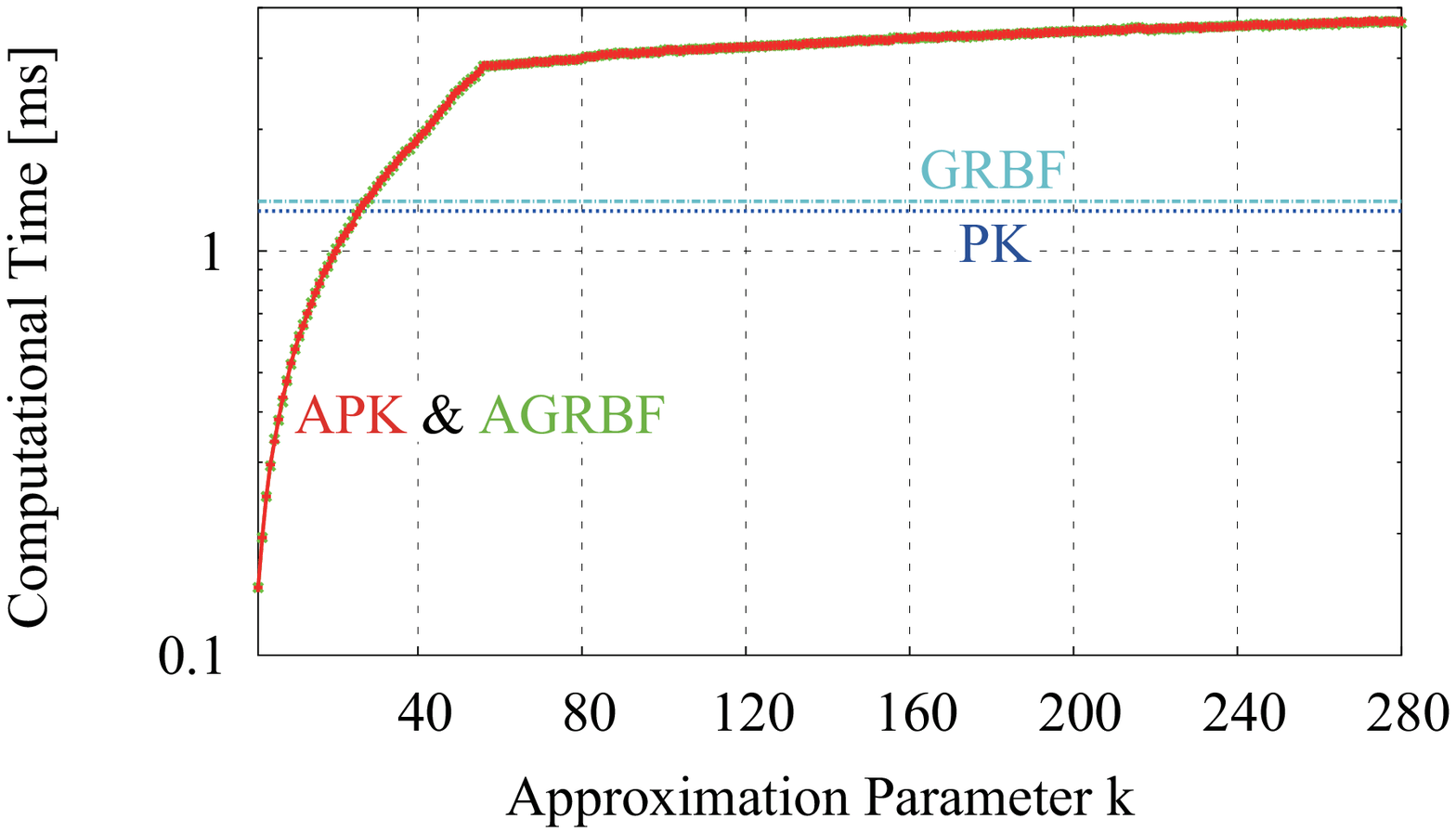}
      \label{fig:ETH80_k_vs_Time}
    }

    \subfloat[Time vs Accuracy (vs NSS methods)]{
      \includegraphics[width=.95\hsize]{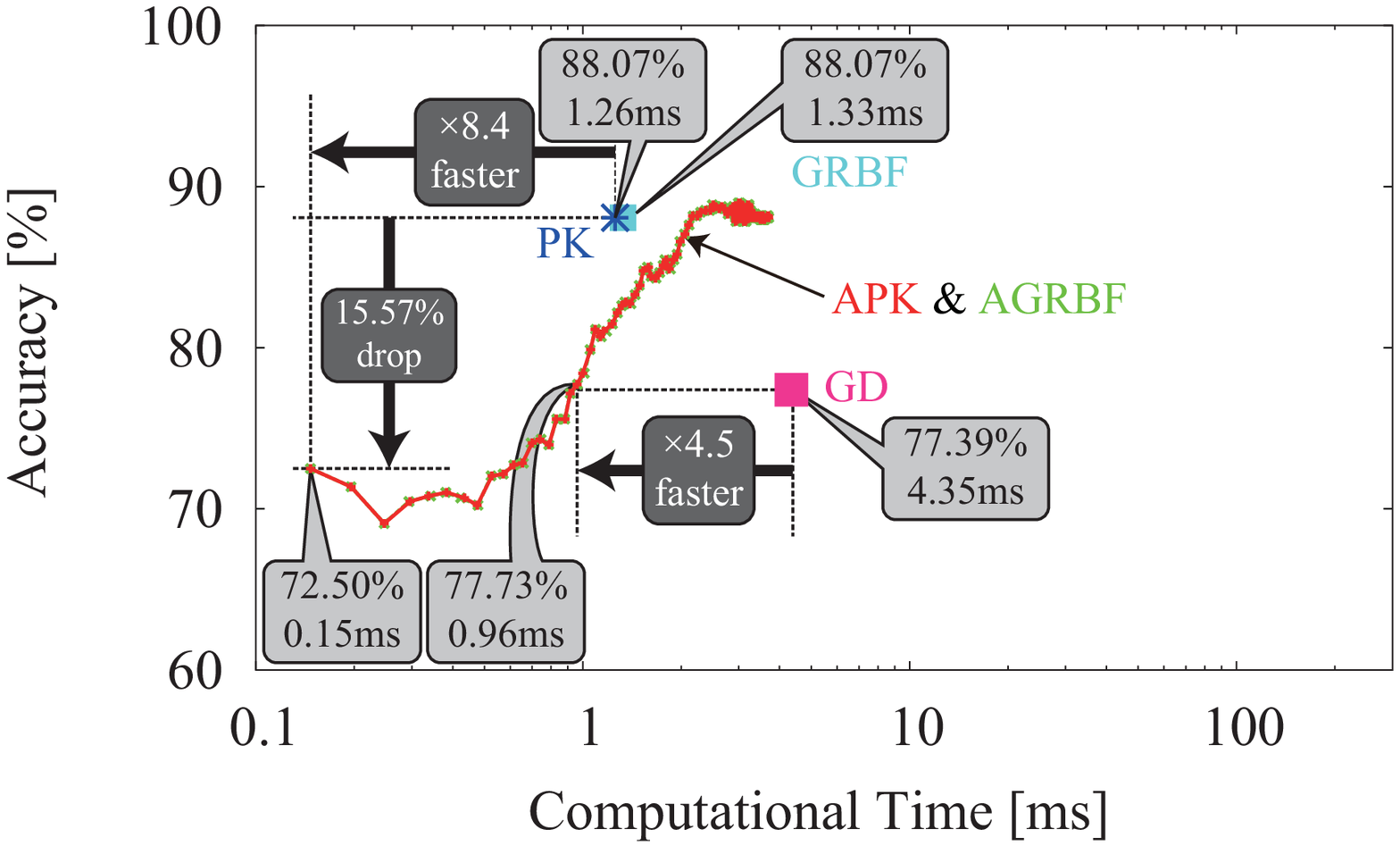}
      \label{fig:ETH80_Time_vs_Acc_NSS}
    }

    \subfloat[Time vs Accuracy (vs ANSS\SSmark\ methods)]{
      \includegraphics[width=.95\hsize]{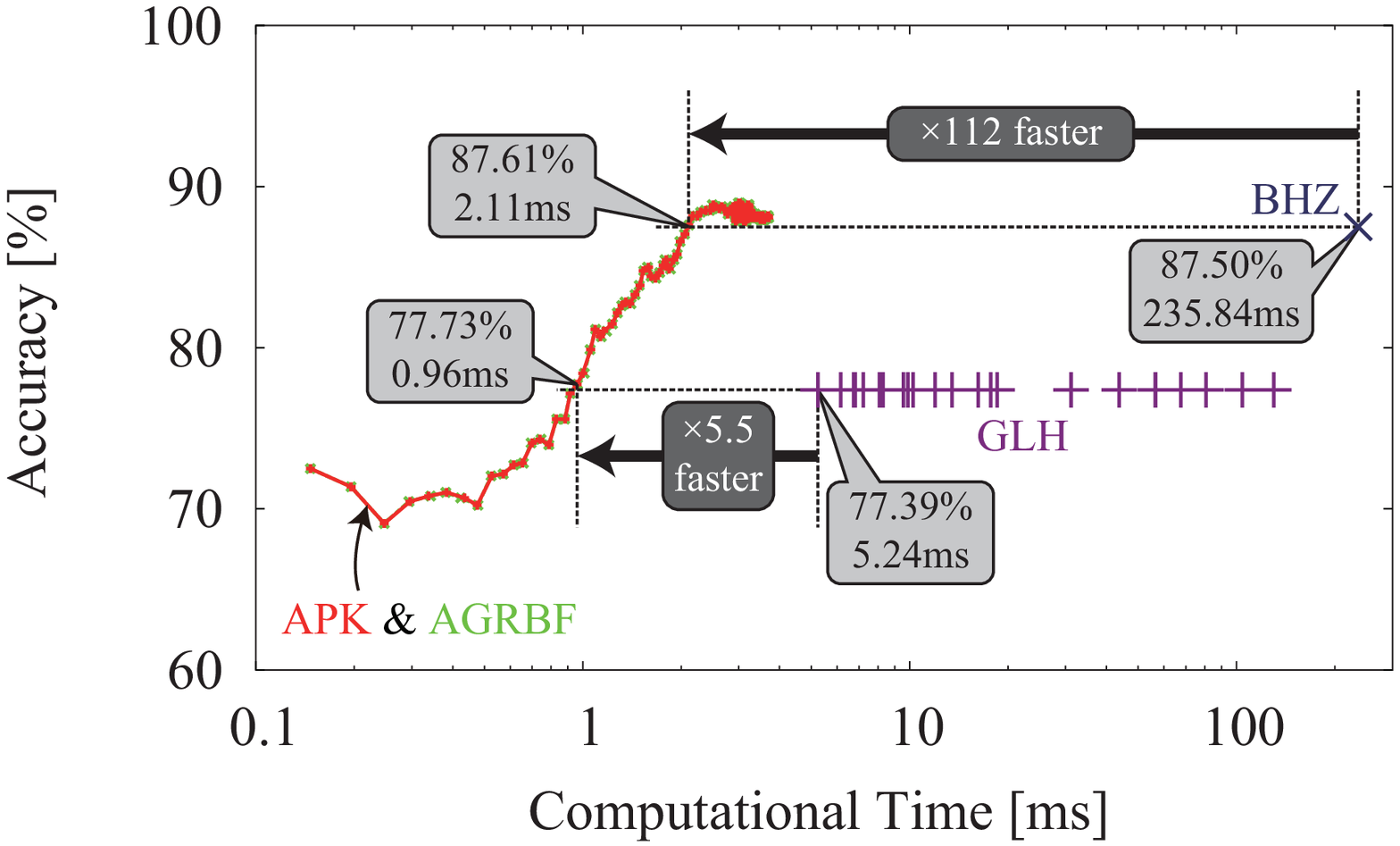}
      \label{fig:ETH80_Time_vs_Acc_ANSS}
    }
    \caption{Experimental results on the object recognition task.}
    \label{fig:ETH80}
  \end{center}
\end{figure}
The recognition results are shown in Fig.~\ref{fig:ETH80}.
In Figs.~\subref*{fig:ETH80_k_vs_Acc} and \subref*{fig:ETH80_k_vs_Time},
the accuracies and computational times of
both proposed methods APK and AGRBF
almost monotonically increased as $k$ of the proposed methods increased.
Their accuracies rapidly increased and
reached the accuracy of PK and GRBF at $k=45$.
At the same $k$,
the computational times of APK and AGRBF became 1.72 and 1.64 times
larger than those of PK and GRBF, respectively,
due to the computational overhead.
In Figs.~\subref*{fig:ETH80_Time_vs_Acc_NSS} and
\subref*{fig:ETH80_Time_vs_Acc_ANSS},
APK was compared with existing NSS and ANSS\SSmark\ methods.
Fig.~\subref*{fig:ETH80_Time_vs_Acc_NSS} shows that
APK was 4.5 times faster than GD
in almost same accuracy,
and
was 8.4 times faster than PK with a 15.57\% loss of accuracy.
Fig.~\subref*{fig:ETH80_Time_vs_Acc_ANSS} shows that
without loss of accuracy,
APK was 112 and 5.5 times faster than BHZ and GLH, respectively.



\subsection{Handwritten Japanese character recognition}
\label{subsec:ETL9B}

We used the handwritten Japanese character dataset ETL9B,
which is the binarized version of the ETL9
released in the 1980s~\cite{Saito_IEICE1985en}.
Sample images are shown in Fig.~\ref{fig:etl9b}.
It contained handwritten Japanese characters of 3036 categories.
Each character category had 200 samples written by 200 subjects.
The first 100 and latter 100 samples per category were
used for training and testing, respectively.

The character images in the ETL9B dataset were
of $64 \times 63$ pixels and
in black and white.
They were nonlinearly normalized to be adjusted to
a $64 \times 64$-pixel grid
using a nonlinear normalization method~\cite{yamada90:nonlinear}.
Then, 1024- and 256-dimensional feature vectors were calculated.
In the 1024-dimensional feature vectors,
$2 \times 2$ image patches were laid out on
a $64 \times 64$-pixel image without overlapping, and
the sum of four pixel values (from 0 to 4) was used as a feature.
Then, 1024-dimensional feature vectors were obtained.
The 256-dimensional feature vectors were created in an almost same
manner.
The only difference was that $4 \times 4$ image patches were used
instead of the $2 \times 2$ image patches.

The dimensionality of subspaces was explored
in the same manner as the object recognition task,
and the best accuracy of PK was achieved with $m=1$
in both 1024- and 256-dimensional feature vectors.
However, with $m=1$, it is not an NSS task.
For the purpose of confirming
the effectiveness of the proposed methods as ANSS\SSmark\  methods,
we selected $m=5$.
Since the number of subspaces in the database was 3036,
the total number of inner products was 75900.
With $k = 7590$,
the proposed methods achieve the same accuracies as
the ones without approximation.

\begin{figure}[tbp]
  \begin{center}
    \subfloat[$k$ vs Accuracy]{
      \includegraphics[width=.95\hsize]{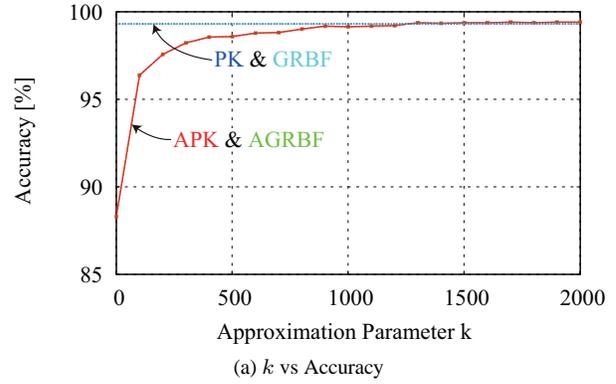}
      \label{fig:ETL9B_16_k_vs_Acc}
    }

    \subfloat[$k$ vs Time]{
      \includegraphics[width=.95\hsize]{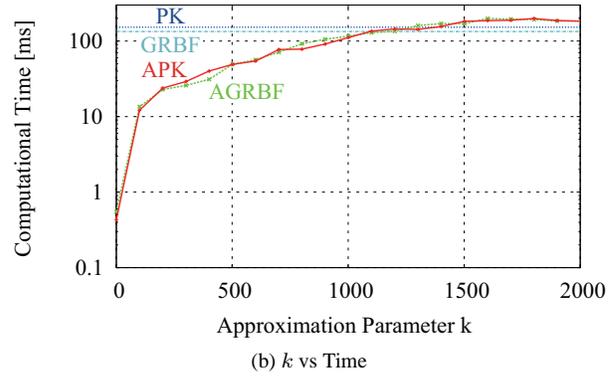}
      \label{fig:ETL9B_16_k_vs_Time}
    }

    \subfloat[Time vs Accuracy (vs NSS methods)]{
      \includegraphics[width=.95\hsize]{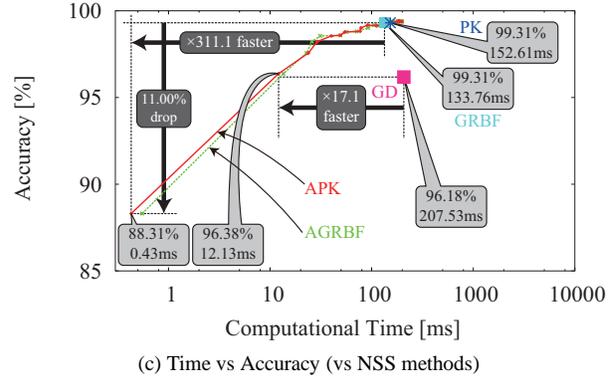}
      \label{fig:ETL9B_16_Time_vs_Acc_NSS}
    }

    \subfloat[Time vs Accuracy (vs ANSS\SSmark\ methods)]{
      \includegraphics[width=.95\hsize]{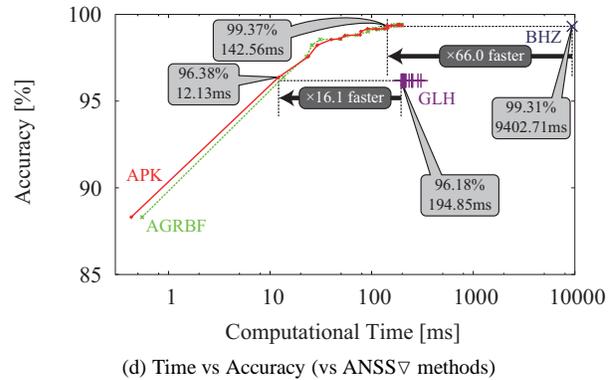}
      \label{fig:ETL9B_16_Time_vs_Acc_ANSS}
    }
    \caption{Experimental results on the character recognition task
      with the 256-dimensional feature vectors.}
    \label{fig:ETL9B_16x16}
  \end{center}
\end{figure}

\begin{figure}[tbp]
  \begin{center}
    \subfloat[$k$ vs Accuracy]{
      \includegraphics[width=.95\hsize]{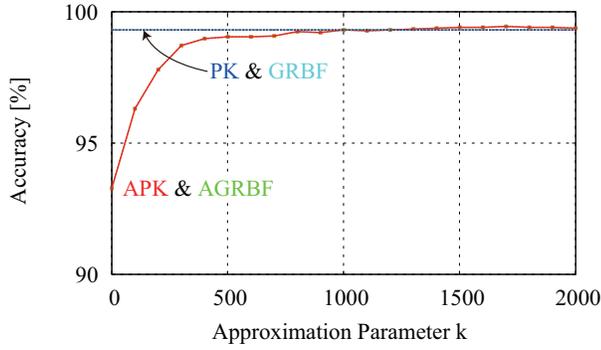}
      \label{fig:ETL9B_32_k_vs_Acc}
    }

    \subfloat[$k$ vs Time]{
      \includegraphics[width=.95\hsize]{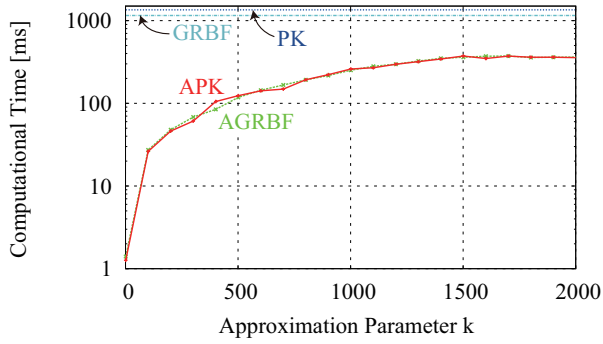}
      \label{fig:ETL9B_32_k_vs_Time}
    }

    \subfloat[Time vs Accuracy (vs NSS methods)]{
      \includegraphics[width=.95\hsize]{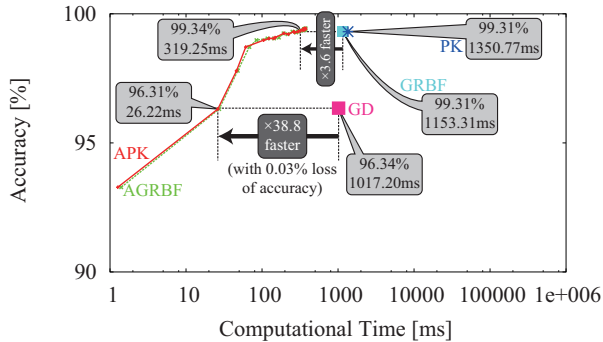}
      \label{fig:ETL9B_32_Time_vs_Acc_NSS}
    }

    \subfloat[Time vs Accuracy (vs ANSS\SSmark\ methods)]{
      \includegraphics[width=.95\hsize]{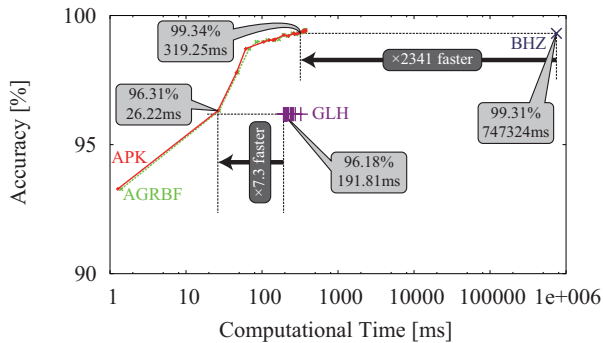}
      \label{fig:ETL9B_32_Time_vs_Acc_ANSS}
    }
    \caption{Experimental results on the character recognition task
      with the 1024-dimensional feature vectors.}
    \label{fig:ETL9B_32x32}
  \end{center}
\end{figure}

The recognition results of 256- and 1024-dimensional feature vectors
are shown in Figs.~\ref{fig:ETL9B_16x16} and
\ref{fig:ETL9B_32x32}, respectively.
Figs.~\subref*{fig:ETL9B_16_k_vs_Acc}, \subref*{fig:ETL9B_16_k_vs_Time},
\subref*{fig:ETL9B_32_k_vs_Acc} and \subref*{fig:ETL9B_32_k_vs_Time}
show the same tendency observed in the object recognition task;
the accuracies and computational times of the proposed
methods monotonically increased as $k$ increased.
With the 256-dimensional feature vectors,
Fig.~\subref*{fig:ETL9B_16_k_vs_Acc} shows that
both APK and AGRBF reached the accuracy of PK and GRBF at $k=1301$.
At the same $k$,
Fig.~\subref*{fig:ETL9B_16_k_vs_Time} shows that
their computational times became 0.93 and 1.2
times of those of PK and GRBF, respectively.
With the 1024-dimensional feature vectors,
Fig.~\subref*{fig:ETL9B_32_k_vs_Acc} shows that
both APK and AGRBF reached the accuracy of PK and GRBF at $k=1201$.
At the same $k$,
Fig.~\subref*{fig:ETL9B_32_k_vs_Time} shows that
their computational times became 0.22 and 0.26
times of those of PK and GRBF, respectively.
In Figs.~\subref*{fig:ETL9B_16_Time_vs_Acc_NSS},
\subref*{fig:ETL9B_16_Time_vs_Acc_ANSS},
\subref*{fig:ETL9B_32_Time_vs_Acc_NSS} and
\subref*{fig:ETL9B_32_Time_vs_Acc_ANSS},
APK was compared with existing NSS and ANSS\SSmark\ methods.
With the 256-dimensional feature vectors,
Fig.~\subref*{fig:ETL9B_16_Time_vs_Acc_NSS} shows that
APK was 17.1 times faster than GD in almost same accuracy,
and
was 311.1 times faster than GRBF with a 11.00\% loss of accuracy.
Fig.~\subref*{fig:ETL9B_16_Time_vs_Acc_ANSS} shows that
without loss of accuracy,
APK was 66.0 and 16.1 times faster than BHZ and GLH, respectively.
With the 1024-dimensional feature vectors,
Fig.~\subref*{fig:ETL9B_32_Time_vs_Acc_NSS} shows that
in almost same accuracy,
APK was 38.8 and 3.6 times faster than GD and GRBF, respectively.
Fig.~\subref*{fig:ETL9B_32_Time_vs_Acc_ANSS} shows that
without loss of accuracy,
APK was 2341 and 7.3 times faster than BHZ and GLH, respectively.
Comparing the results with the 256- and 1024-dimensional feature vectors,
the advantage of the proposed methods to the other methods was larger
with the 1024-dimensional feature vectors. 
This supports the effectiveness of the proposed methods.

\section{Conclusion}

In this paper, we presented a scalable solution for
the approximate nearest subspace search problem.
The proposed methods are computationally efficient
in a large-scale problem,
with regard to both the number of subspaces and
the dimensionality of the feature space.
The key idea is to decompose a distance calculation
in the Grassman manifold into multiple distance calculations
in the Euclidean space.
This makes it possible to efficiently approximate the distance
calculation even in the difficult conditions.
In the experiment with 3036 subspaces in the 1024-dimensional space,
we confirmed that one of the proposed methods was
7.3 times faster than the previous state-of-the-art
without loss of accuracy.

Future work includes an extension of the proposed methods
to cope with subspaces of variable dimensionalities.
The limitation of the dimensionalities of subspaces
comes from the definition of the Grassmannian and
all subspace distances defined on it have the same problem.
This problem may be solved by using latest results
such as \cite{Ye_arXiv2014}.

\section*{Acknowledgment}
This work is partially supported by JST CREST project, and
JSPS KAKENHI \#25240028.

\clearpage

{\small
\bibliographystyle{IEEEtran}
\bibliography{paper}
}

\end{document}